\documentclass[english]{article}
\usepackage[T1]{fontenc}
\usepackage[latin9]{inputenc}
\usepackage{amssymb}
\usepackage{graphicx}

\makeatletter

\providecommand{\tabularnewline}{\\}

\makeatother

\usepackage{babel}
\usepackage[nonatbib,final]{nips_2016}

\begin{document}

\title{Interpretation of Prediction Models Using the Input Gradient}
\author{
  Yotam Hechtlinger \\
  Department of Statistics\\
  Carnegie Mellon University\\
  \texttt{yhechtli@stat.cmu.edu} \\
}

\maketitle

\section*{Abstract}

State of the art machine learning algorithms are highly optimized to provide the optimal prediction possible, naturally resulting in complex models. While these models often outperform simpler more interpretable models by order of magnitudes, in terms of understanding the way the model functions, we are often facing a ``black box''. 

In this paper we suggest a simple method to interpret the behavior of any predictive model, both for regression and classification. Given a particular model, the information required to interpret it can be obtained by studying the partial derivatives of the model with respect
to the input. We exemplify this insight by interpreting convolutional and multi-layer neural networks in the field of natural language processing.% and image classification. 

\section{Introduction}

The property of interpretability of a model can be defined in several
ways. In this paper, we are interested in the following property:
given a prediction model $\hat{f}:X\rightarrow Y$, where $X$ is
the input space and $Y$ is the output space, we would like interpret
which variables in $X$ affect the prediction of $\hat{f}$ and in
which way. When this property is achieved we say the model $\hat{f}$
is interpretable. 

In linear regression interpreting a given model relies on the model
parameters. Since the model is of the form of $\hat{f}\left(x\right)=\hat{\beta}^{T}x$,
there is a $1$--$1$ correspondence between the parameters and the coordinate of the variables in the input space.
Therefore, the effect of a given variable $x_{k}$ on the model $\hat{f}$,
can be evaluated through the value of $\hat{\beta_{k}}$. When $\hat{\beta_{k}}\gg0$
we may say $x_{k}$ has a significant positive effect on $\hat{f}\left(x\right)$.
When $\hat{\beta_{k}}\approx0$, we may say that $x_{k}$ is irrelevant
to the prediction of $y$, and it may get selected out of the model.
Intuitively it is often also explained in terms of a slope: One unit
increase in $x_{k}$ would increase the prediction by the amount of
$\hat{\beta_{k}}$. 

This approach of inferring variables importance through the parameters,
has been thoroughly studied by statisticians, since it is fundamental
to the field of statistical inference. It has been expanded from linear
regression to generalized linear models, and can also be applied to
other algorithms which rely on the inner product of the
data and the variables, such as SVM. As models become more complex
there is no longer a $1$--$1$ relation between the parameters and the
variables. For example, in neural networks, the number of parameters
will be magnitude larger than the number of variables, and the relation
between a variable and specific set of parameters is often intractable. 

The key observation and main point of this paper is that it is possible
to interpret the model without using the parameters. Essentially it
is possible to interpret any $\hat{f}$ by studying the partial derivatives
of $\hat{f}$ with respect to the input variables $x\in X$. The intuition is
straightforward. Suppose that $\forall x\in X$, it happens that $\frac{\partial\hat{f}}{\partial x_{k}}\left(x\right)=0$
for the variable $x_{k}$ , then effectively $x_{k}$ doesn't influence
$\hat{f}$ at all. This is also true the other way around. If $\forall x\in X$,
$\left|\frac{\partial\hat{f}}{\partial x_{k}}\left(x\right)\right|\gg0$,
we (intuitively) learn that $x_{k}$ is important to the model prediction. 

For the particular case of linear regression, when $\hat{f}\left(x\right)=\hat{\beta}^{T}x$,
the two approaches coincide since $\frac{\partial\hat{\beta}^{T}x}{\partial x_{k}}=\hat{\beta}_{j}$, and the parameters themselves are the partial derivatives. The major
advantage of the suggested perspective is that it can be generalized to any given prediction model $\hat{f}$, both for regression and classification problems. 

In the paper we show how it is possible to interpret complex models, specifically, convolutional and multi-layer neural networks using data from the field of natural language processing.

\section{Literature Review}

The idea of evaluating the performance of the model through the partial derivatives of the input variables is seemingly trivial, but we could not locate examples in the literature in which it has been done for interpretation purposes. The most relevant reference we could find is Google DeepDream \cite{deepDream2015}, which uses GoogLeNet convolutional neural net  \cite{szegedy2015going}. They calculate the gradients of a given input image and perform a gradient ascent on the input space to maximize particular neurons or layers activations. This provide a visualization tool to enable the interpretation of the convolutional neural network. 

A recent paper by Riberio et al \cite{ribeiro2016should} approach the problem of interpreting a given model from a different direction, with a new technique named LIME aimed to interpret any prediction model. In the paper the authors suggest to interpret the performance of $\hat{f}$ at a particular point $x\in X$, by approximating $\hat{f}$ locally around $x$ with a model $g$ which is simple and interpretable, such as sparse linear regression (``Lasso'') or a decision tree. 

\section{Method and Notation}
This paper consider a prediction model $\hat{f}:X\rightarrow Y$, learned from observations of the input space $X$ and the output space $Y$. This is the general framework used in supervised learning problems, including both regression, when $Y$ is continuous and classification, when $Y$ is categorical. $X$ is a high dimensional feature space, with $p$ different features, $X_{1},\ldots,X_{p}$, where $X_{k}$ can also be discrete or categorical. We denote the training set as $X_{train}$ and testing as $X_{test}$. 

We propose to study the gradient values of $\hat{f}$ with respect to the features. Formally, an observation $x\in X$, is in the form of $x=\left(x_{1},\ldots,x_{p}\right)$, where $x_{k}$ is the value of the $k$-th variable (or feature). We will study the gradient $\nabla\hat{f}(x)=\left(\frac{\partial f}{\partial x_{1}},\ldots,\frac{\partial f}{\partial x_{p}}\right)$.  The $k$-th partial derivative will be denoted as $g_{k}\left(x\right)\equiv\frac{\partial\hat{f}}{\partial x_{k}}\left(x\right)$. We will also use the average of the partial derivatives in the test set,
$\bar{g}_{k}=\frac{\sum_{x\in X_{test}}g_{k}\left(x\right)}{\left|X_{test}\right|}$,
denoted in vector form as $\widetilde{g}=\left(\bar{g}_{1},\ldots,\bar{g}_{p}\right)$.

The computation of the gradients in neural networks, and often in other models, can be derived explicitly using the backpropagation and the chain rule. This is what we have done in this paper. In other cases it may be possible to estimate it numerically. For a small $h$, $\frac{\partial\hat{f}}{\partial x_{k}}\approx\frac{\hat{f}\left(x_{1},\ldots,x_{k}+h,\ldots,x_{p}\right)-\hat{f}\left(x_{1},\ldots,x_{k},\ldots,x_{p}\right)}{h}$. When $x_{k}$ is continuous, this is quickly computable, and only requires two forward passes per coordinate, since $\hat{f}$ is given. In ~\ref{subsec:CNN} and ~\ref{subsec:BoW} we will discuss cases when $x_{k}$ is categorical and $\hat{f}$ not differentiable.

In addition, the gradient, $\nabla{\hat{f}}$, is a function defined globally on the entire space. When it can not be analytically derived, as in most cases, it has to be evaluated at multiple points locally. In all of the examples we trained a model on the training set, and evaluated $\nabla{\hat{f}}$ on the test set. This is reasonable due to the standard assumption that the training and test sets are an $i.i.d$ sample from the population, but this course of action should be evaluated on a case-by-case basis.

\section{Examples}

%In order to demonstrate the effectiveness of the gradients for interpretation purposes, we present $3$ examples from the field of text analysis and image classification of complex models being interpreted using the gradients. 

In order to demonstrate the effectiveness of the gradients for interpretation purposes, we present $2$ examples from the field of natural language processing in which complex models are being interpreted using the gradients. The examples are using the IMDB Large Movie Review Dataset \cite{maas2011learning}. The dataset contains $50,000$ different movie reviews, equally partitioned to train and test sets, with $Y$ being the labeled sentiment of the review - either positive and negative. %For the image classification example we are using the well studied MNIST dataset [[CITE]] for the problem of hand digits classification. 

Implementation has been done using Keras \cite{chollet2015keras} and Theano \cite{2016arXiv160502688short}. Since Theano is doing symbolic differentiation, calculating the gradients efficiently was a simple short function call. Source code enabling replication of the results is available on the author website. 

%Text is an interesting data to study for the problem of interpretability, because (1) it is highly interpretable to humans and (2) the data is in the form of a discrete dictionary with sequential effect, requiring complex modeling. In this test case we used the IMDB Large Movie Review Dataset {[}CITE{]} which contains $50,000$ different movie reviews, equally partitioned to train and test sets. The target is to predict the review sentiment directly from the text, with half of the reviews labeled as positive while the other half negative. 

\subsection{Example $1$: Convolutional Neural Networks following Word Embedding}
\label{subsec:CNN}
There are two major challenges with textual data. First the words are categorical and part of a dictionary. Second, significant information can be learned through the sequential order of the sentence. In recent years many state-of-the-art benchmarks in textual problems are being achieved using convolutional neural networks (CNN) models. CNN model address the first challenge by embedding the dictionary in a high dimensional metric space, and the second by applying convolutions to capture the sequential component of the data. See \cite{collobert2011natural} for model details and \cite{kim2014convolutional} for performance review. 

%The model has two crucial components. First the words dictionary is being embedded to high dimensional vector, transforming the text from a sequential list of words to a sequential list of vectors. Then a one dimensional convolutions are being applied on the sequence according to the sentence order, extracting features from the text, followed by pooling layer and linear classifier. 

Interpreting a CNN model with textual data is not a trivial task because the original sentence is hardly intractable after the embedding and the convolutions. Nevertheless, in this example we demonstrate how it is achievable by using the gradients to interpret the model.  

To train a CNN on the IMDB dataset, we thresholded the review length to $400$, and the dictionary size to the $5000$ most popular words. For a sentence $x=\left(x_{1},\ldots x_{400}\right)$, where $x_{k}$ is the $k$-th word, the embedding maps each word $x_{k}\mapsto z_{k}\in\mathbb{R}^{50}$ and transfer $x\mapsto z\in\mathbb{R}^{400\times50}$. After the embedding, we apply $250$ convolutions on $z$ with length $3$ (words), depth $50$ (embedding dimension), followed by max pooling and a linear classifier. This model yields a $89.2\%$ accuracy, not far from the state-of-the-art (topping $90\%$ has been done a handful of times and is considered a significant improvement~\cite{johnson2016supervised}).

To interpret this complex model, we turn to gradients. Since the words themselves are categorical, the prediction function is not differentiable with respect to the words. We therefore use the chain rule to calculate for every word the partial derivative of the model with respect to the embedding vector. Thus, for the $k$-th word the gradient is $\frac{\partial\hat{f}}{\partial z_{k}}\in\mathbb{R}^{50}$. This provides for each sentence $400$ words gradients, which we rank according to their $\ell_{2}$ norm in order to assess which ones are the most influential. The intuition toward taking the highest norm is due to the fact that the norm of the gradient is a proxy for the "magnitude" of the slope of the graph in the direction of the gradient vector, and the steepest slope will identify the word effecting the prediction the most. 

\begin{table}
\begin{centering}
\begin{tabular}{|l|l|l|}
\hline 
Ex.1 - Positive Sentiment & Ex. 2 - Negative Sentiment & Ex. 3 - Negative Sentiment\tabularnewline
\hline 
\textit{fantastic film total} & \textit{unfortunately they suffer} & \textit{bad as hell}\tabularnewline
\textit{moving highly entertaining} & \textit{dull after five} & \textit{badly stop motion}\tabularnewline
\textit{unconvincing i can't} & \textit{painfully dull after} & \textit{horrible you won't}\tabularnewline
\textit{highly entertaining references} & \textit{becomes painfully dull} & \textit{was bad as}\tabularnewline
\hline 
\end{tabular}
\par\end{centering}
\caption{The four most relevant expressions according to the norm of the gradients
in the CNN model ordered from top to bottom. The first word in the sentence is the one selected according to the ordering of the gradient, while the following two are the rest of the expression being forwarded into a length $3$ convolutions. It can be seen from Ex.2 that the word \textit{"dull"}
is effecting the prediction through all $3$ locations of the convolution.}
\label{tab:CNN results}
\end{table}

Table~\ref{tab:CNN results} shows the expressions for which the gradients norm are the largest for several reviews in the test set. Since the convolution is being applied on length of $3$ words we present the expressions activating the convolution, where the first word is the one with the highest gradient norm. The highest ranked expressions are highly interpretable, and most examples are self explanatory, such as \textit{"fantastic film total"} or \textit{"lack of credibility"}. Some expressions were activated by the words following the selected one, such as \textit{"ape was outstanding"}, in which the word \textit{"ape"} had the largest gradient in the sentence, but probably explained as influential by the \textit{"was outstanding"} expression. 

\subsection{Example $2$: Bag of Words Model}
\label{subsec:BoW}

In the previous example we have used the gradients to interpret the local information about a particular sentence. In this example we will use the gradients globally in order to assess which words affect the classifier the most, and draw insight on the classifier functionality. To do so we represent the sentences with a simplified version of the Bag of Words (BoW) model. 

A sentence $x\in X$ is represented as a binary vector in the length of the dictionary $D$, in which $x_{d}=1$ if the $d$ word occur in the sentence, and $0$ otherwise. In this model the data representation is simpler than the word embedding used in example $1$, since it discard the grammar and sequential information from the original text. In general it tends to under-perform the word embedding representations, but it is easier to interpret, since now there is $1$--$1$ correspondence between the features and the words. Following the BoW data representation we fit a fully connected neural network with $2$ hidden layers to perform the classification, resulting with a $85.4\%$ model accuracy. 

Next, for all $x\in X_{test}$ the partial derivatives are evaluated, and $\widetilde{g}=\left(\bar{g}_{1},\ldots,\bar{g}_{D}\right)$, the vector of the averages of the partial derivatives, is calculated. $\widetilde{g}$ provides a single global estimator of the gradient function across the input space, which can be used to rank the words according to their influence on the model. As can be seen in Table~\ref{tab:BoW results}, the top positive and negative words according to the values of $\widetilde{g}$ are highly interpretable, and include words such as \textit{"excellent"} and \textit{"great"} for positive sentiment and \textit{"worst"} and \textit{"waste"} for negative sentiment. 

Furthermore, $\widetilde{g}$ can also be used for approximating the prediction function $\hat{f}$, since the features are binary. By classifying $x_{new}$ as $1$ if $\left\langle \widetilde{g},x_{new}\right\rangle >0$ and $0$ otherwise, we get a classifier that agrees with the prediction function $\hat{f}$ on $\frac{24907}{25000}\approx99.6\%$ of the observations in the test set. This surprising result provide further insight on $\hat{f}$, specifically that (1) the decision boundary of $\hat{f}$ is closely defined by the hyper-planes created by $\widetilde{g}$, therefore the learned $\hat{f}$ is linear although the model is much more complex; (2) $\bar{g}_{d}$ can be used as a trustworthy estimation of influence of a given word.

There is a last nuance about this example that should be discussed. Although the function is not differentiable because $x_{j}\in\left\{ 0,1\right\}$, the derivatives were obtained as if it were. Effectively it means that the derivative is being evaluated as if $x_{j}\in\left(-\epsilon,1+\epsilon\right)$. This approach can be extended to general categorical data with the use of indicators as features. 

\begin{table}
\begin{centering}
\begin{tabular}{ll|ll}
\multicolumn{2}{c}{Positive} & \multicolumn{2}{c}{Negative}\tabularnewline
\hline 
\textit{excellent} & $\left(0.096\right)$ & \textit{worst} & $\left(-0.160\right)$\tabularnewline
\textit{great} & $\left(0.091\right)$ & \textit{waste} & $\left(-0.117\right)$\tabularnewline
\textit{perfect} & $\left(0.085\right)$ & \textit{boring} & $\left(-0.105\right)$\tabularnewline
\textit{amazing} & $\left(0.082\right)$ & \textit{awful} & $\left(-0.103\right)$\tabularnewline
\textit{wonderful} & $\left(0.080\right)$ & \textit{bad} & $\left(-0.097\right)$\tabularnewline
\hline 
\end{tabular}
\par\end{centering}
\caption{Most influential words for each sentiment to the prediction of the BoW
model, ordered from top to bottom. In parenthesis we show the average of the partial derivatives over the test set. It is approximately the effect the specific word has on the model prediction. }
\label{tab:BoW results}
\end{table}

\section{Conclusion}

In this paper we have demonstrated the usefulness of the model gradient vector with respect to the input, as an important quantity to use when studying the behavior of a given model. The gradient vector indicates which features affect the prediction the most, thus it can be used to draw insights and to interpret the input features in complex models quickly and efficiently. We have also showed that the gradients contains enough information to draw global conclusions on the behavior of the classifier,
including the linear structure of the model, and create a linear approximation of the function. 

\section{Acknowledgments}
We would like to thank Ruslan Salakhutdinov and Alessandro Rinaldo for suggestions, insights and remarks that has greatly improved the quality of the paper.

\bibliography{reference}
\bibliographystyle{plain}

\end{document}